%% file: main.tex
\documentclass[10pt,twocolumn,letterpaper]{article}

\usepackage[pagenumbers]{cvpr} 

\input{preamble}

\definecolor{cvprblue}{rgb}{0.21,0.49,0.74}
\usepackage[pagebackref,breaklinks,colorlinks,citecolor=cvprblue]{hyperref}

\usepackage{algorithm}
\usepackage{algorithmic}
\usepackage{utfsym}

\title{Prompt-Driven Dynamic Object-Centric Learning for Single Domain Generalization}

\author{Deng Li\textsuperscript{1}, Aming Wu\textsuperscript{2}, Yaowei Wang\textsuperscript{3}, Yahong Han\textsuperscript{1,\thanks{Corresponding author.}}\\
\textsuperscript{1}College of Intelligence and Computing, Tianjin University, China\\
\textsuperscript{2}School of Electronic Engineering, Xidian University, China\\
\textsuperscript{3}Peng Cheng Laboratory, China\\
{\tt\small lideng@tju.edu.cn, amwu@xidian.edu.cn, wangyw@pcl.ac.cn, yahong@tju.edu.cn}
}

\begin{document}
\maketitle
\input{0_abstract}
\input{1_intro}

\input{2_relatedworks}

\input{3_method}

\input{4_experiments}
{
    \small
    \bibliographystyle{ieeenat_fullname}
    \bibliography{main}
}


\end{document}

%% file: preamble.tex
\usepackage[dvipsnames]{xcolor}


%% file: 0_abstract.tex
\begin{abstract}

Single-domain generalization aims to learn a model from single source domain data to achieve generalized performance on other unseen target domains. Existing works primarily focus on improving the generalization ability of static networks. However, static networks are unable to dynamically adapt to the diverse variations in different image scenes, leading to limited generalization capability. Different scenes exhibit varying levels of complexity, and the complexity of images further varies significantly in cross-domain scenarios. In this paper, we propose a dynamic object-centric perception network based on prompt learning, aiming to adapt to the variations in image complexity. Specifically, we propose an object-centric gating module based on prompt learning to focus attention on the object-centric features guided by the various scene prompts. Then, with the object-centric gating masks, the dynamic selective module dynamically selects highly correlated feature regions in both spatial and channel dimensions enabling the model to adaptively perceive object-centric relevant features, thereby enhancing the generalization capability. Extensive experiments were conducted on single-domain generalization tasks in image classification and object detection. The experimental results demonstrate that our approach outperforms state-of-the-art methods, which validates the effectiveness and generally of our proposed method.

\end{abstract}

%% file: 1_intro.tex
\section{Introduction}
\label{sec:intro}
 
Recently, deep learning visual models have achieved rapid development \cite{he2016deep, rawat2017deep, zou2023object, carion2020end}. These methods are based on the assumption that the training and testing data share a similar distribution. However, in practical applications, the training and testing data are often not drawn from the same distribution. Deep learning models often exhibit poor generalization performance when tested on unseen or out-of-distribution datasets. The main reason behind this is domain shift \cite{sugiyama2006mixture}, where the distribution of the testing data significantly differs from that of the training data. To mitigate the impact of domain shift, several approaches have been proposed, such as domain adaptation \cite{murez2018image,sankaranarayanan2018generate} and domain generalization \cite{muandet2013domain,ghifary2015domain, carlucci2019domain} methods. 
Domain adaptation methods typically require the inclusion of unlabeled target domain images during the model training phase. Multiple domain generalization methods aim to mitigate the domain shift by combining data from multiple training domains to some extent. However, both of these approaches have limitations due to the expensive data acquisition and data privacy. As a result, these methods may not be applicable in all scenarios.

\begin{figure}[t]
\vspace{-0.5em}
\begin{center}
\includegraphics[width=1 \linewidth]{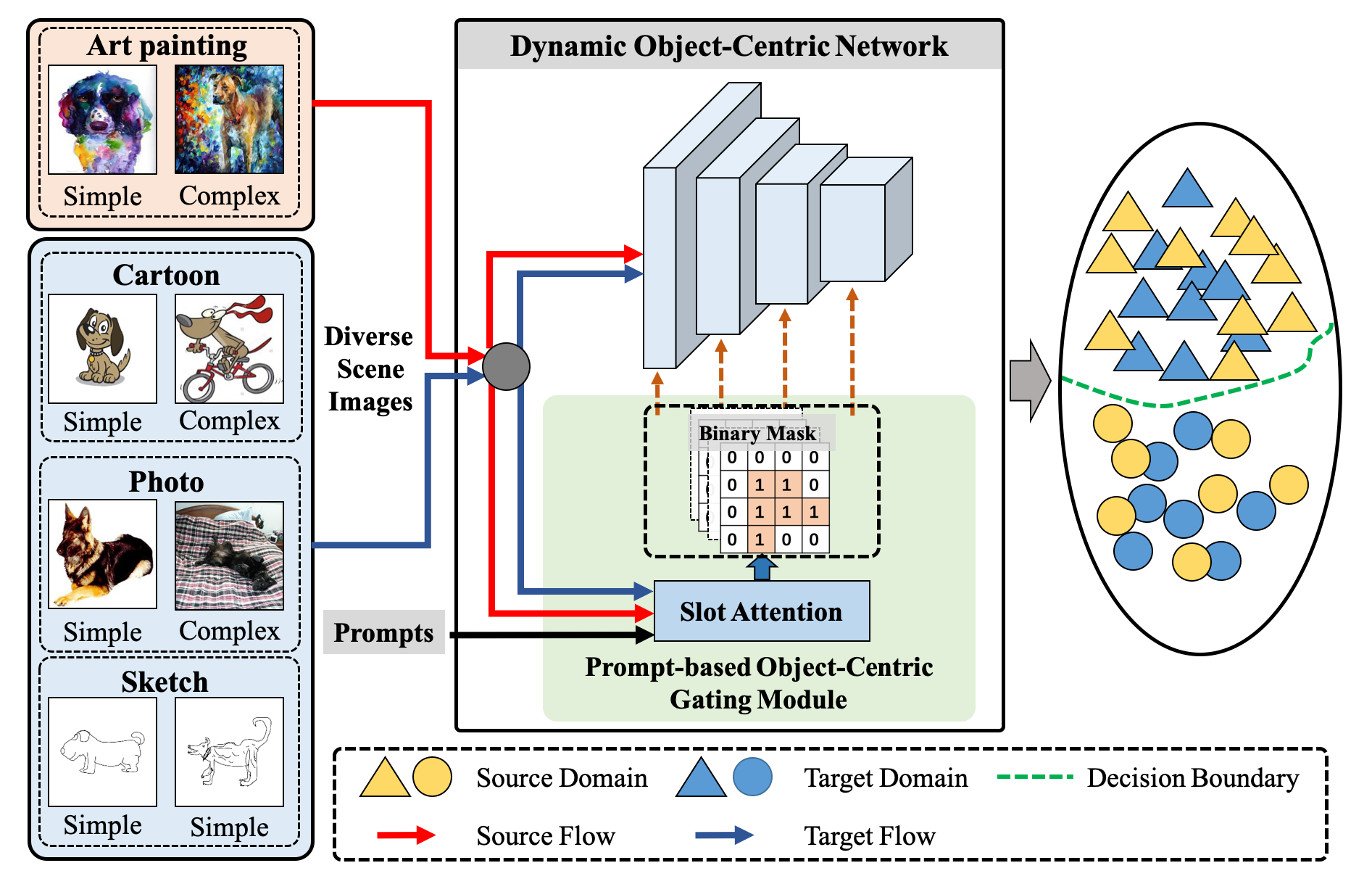}
\end{center}
\caption{Illustration of dynamic object-centric learning via prompts for single domain generalization. 
Object-centric features capture the essential information related to individual objects.
Incorporating the given scene prompts to dynamically optimize the extraction of object-centric features is beneficial for improving the generalization performance of models.
   }
\label{dynamic_sg}
\end{figure}

Single-domain generalization aims to train a model on a single domain and generalize its performance to diverse unseen target domains \cite{volpi2018generalizing}. This learning paradigm poses significant challenges due to the fact that the model is trained only on a single source domain and the target domains are inaccessible during training. 
Existing approaches for single-domain generalization primarily focus on two main methods: data augmentation \cite{qiao2020learning,volpi2018generalizing,volpi2019addressing,zhao2020maximum} and feature disentanglement \cite{wu2022single}. 
While the methods above have made positive contributions to alleviate domain shift in single-domain generalization tasks, they mainly focus on static networks. 
Static networks are unable to dynamically adapt to the diverse variations in different image scenes, which limits the representation power of the models. 
Dynamic networks \cite{han2021dynamic} dynamically adjust the structure or parameters to adapt the characteristics of the input data, expanding the parameter space, and leading to more effective learning. This adaptability enables the model to capture complex patterns and variations in the data and improve the generalization performance.

In the visual tasks, each image may have its own unique characteristics, such as variations in lighting conditions, object appearances, or scene structures, which can result in variations in data complexity. The complexity of data varies across different domains, and consequently, the required network complexity between different images may differ. 
Object-centric representations are robust to variations in appearance, context, or scene complexity, which enables the model to generalize well to unseen or novel samples. 
Considering the above factors, we propose a dynamic object-centric learning approach for single-domain generalization as shown in Figure ~\ref{dynamic_sg}. 
Specifically, a prompt-based object-centric gating module is designed to perceive object-centric features of objects, leveraging the multi-modal feature representation capabilities of the visual-language pre-trained CLIP \cite{radford2021learning} model, and the prompts that describe different domain scenes guide the learning of the dynamic gating decision for different domains. Furthermore, we proposed a Slot-Attention multimodal fusion module to fuse the linguistic features and visual features and then extract effective object-centric representations. With learned object-centric gating decisions, we selectively connect the features of the network in both spatial and channel dimensions.
We validated the effectiveness of our proposed method on different visual tasks of varying complexity, including image classification and object detection.

The main contributions of this work can be summarized as follows:

(1) To address the issue of insufficient generalization ability of single-domain generalization tasks, we propose a dynamic object-centric learning framework to enhance the generalization capability.

(2) We propose an object-centric gating module based on prompt learning which leverages the textual descriptions of various scenes to guide the learning of the gating decision for different domains. Additionally, we introduce a Slot Attention multi-modal fusion module to extract effective object-centric representations.

(3)  Extensive experiments conducted on image classification and object detection tasks of varying complexities validate the effectiveness and generality of the proposed method.

%% file: 2_relatedworks.tex
\section{Related Works}
\label{sec:formatting}

\subsection{Single Domain Generalization}

Single-domain generalization tasks aim to train on a single source domain and generalize to unseen target domains. Existing methods can be divided into two categories: data or feature augmentation and learning domain-invariant features. The data augmentation method aims to generate some out-of-distribution samples at the data level or feature level. In particular, some works \cite{qiao2020learning,volpi2018generalizing,volpi2019addressing,zhao2020maximum} show that the method of adversarial domain augmentation can effectively improve the generalization ability and robustness of the model by synthesizing virtual images during the training process. 
CLIP-Gap \cite{vidit2023clip} utilizes the joint representation space of visual and textual features in the pre-trained multimodal CLIP model to learn the feature shift between the visual and textual descriptions of the target domain. 
\cite{wang2021learning} explores improving generalization capabilities by alternating diverse sample generation and discriminative style-invariant representation learning. Domain-invariant feature learning methods aim to learn feature representations that are invariant to domain variations from the source domain data. 
Wu et al. \cite{wu2022single} proposed a method that disentanglements features into domain-specific and domain-invariant components, and then uses the domain-invariant features as teacher feature representations to enhance the generalization capability of the detection model through self-distillation.

Due to the limitations imposed by the diversity of training data from a single source domain, static networks are prone to overfitting during training.  Different from the above methods, considering that the dynamic network dynamically adjusts the network structure according to the input data, expanding the parameter space of the model and improving the representation power. 
We propose a prompt-based dynamic network single-domain generalization method, which guides the learning of the dynamic gating decision with various domain descriptive text.

\subsection{Dynamic Networks}

\begin{figure*}[]
\vspace{-1.5em}
\begin{center}
\includegraphics[width=0.88 \linewidth]{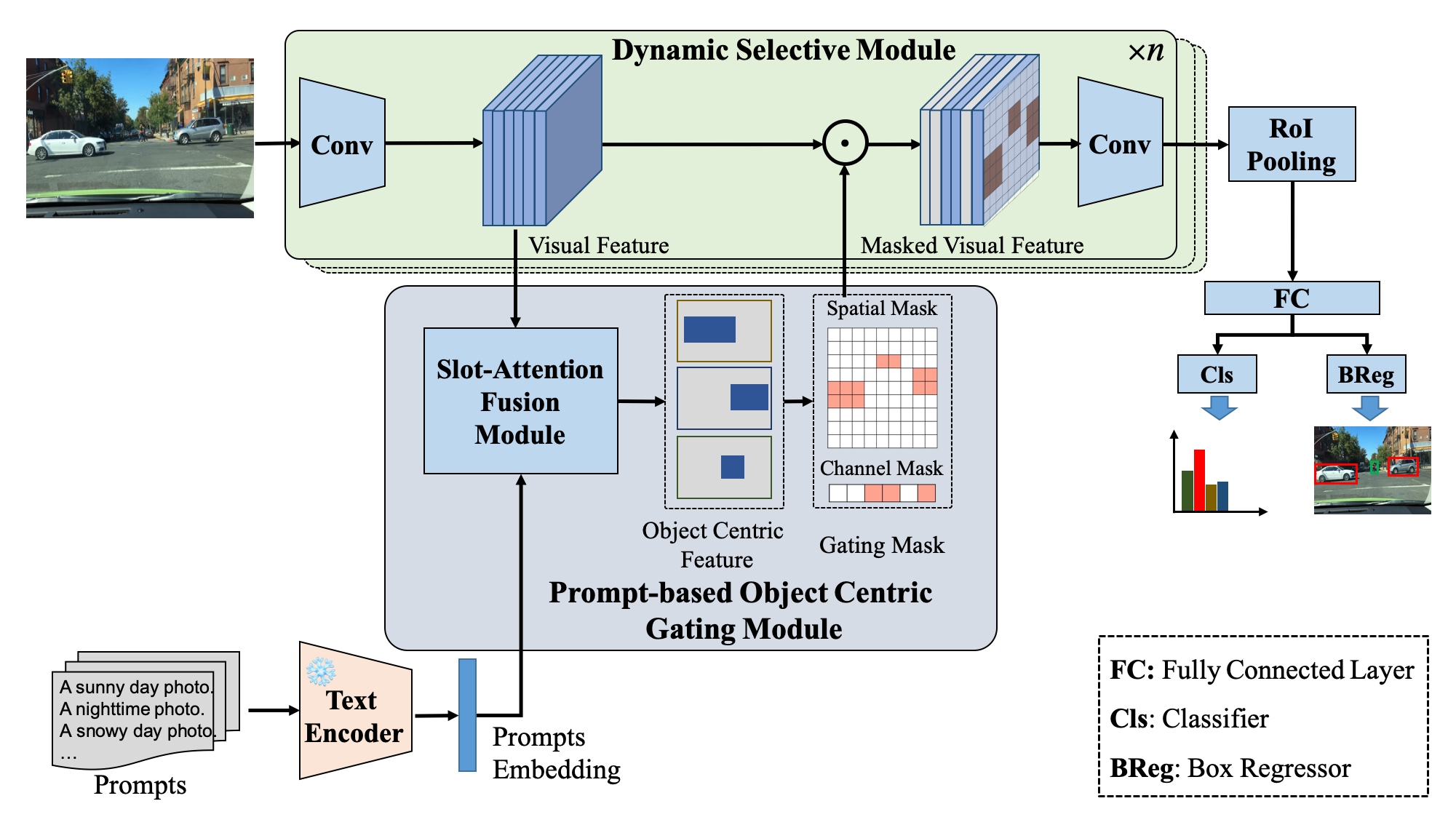}
\end{center}
\caption{Illustration of our proposed prompt-based dynamic object-centric learning network for single domain generalization. 
This method mainly includes a prompt-based object-centric gating module and a dynamic selective module. First, the Slot Attention multimodal fusion module extracts object-centric features and leverages the various scene prompts to guide the object-centric gating mask learning for the input from different scenes. Next, the gating mask is used to dynamically select the relevant object-centric features to improve the generalization ability.
   }
\label{arch}
\end{figure*}

Dynamic networks adaptively adjust their network structure based on input data to perform inference on different input data. These methods can make decisions based on different criteria to select different sub-networks for computational execution. The existing main dynamic network methods can be divided into two categories: early exit and gating function-based methods. BranchyNet \cite{teerapittayanon2016branchynet} and MSDNet \cite{huang2017multi} employ confidence-based criteria to explore early exit methods, which divide the model into multiple stages and handle simpler inputs that require fewer complex stages in the network. 
Some methods, such as BlockDrop \cite{wu2018blockdrop}, GaterNet \cite{chen2019you}, and SBNet \cite{ren2018sbnet}, utilize strategy networks or learn dynamic decisions based on gate functions. 
SkippNet \cite{wang2018skipnet}and ConvNet-AIG \cite{veit2018convolutional} dynamically skip the block module in the residual network \cite{he2016deep} based on input features, while this coarse-grained approach results in considerable accuracy loss. CGNet \cite{hua2019channel} and PGNet \cite{zhang2020precision} take advantage of the sparsity of spatial features to achieve different output activations for the input feature maps. 

The gating function-based method has significant versatility and applicability and can be applied to different aspects of the network. We build an object-centric gating module based on Slot Attention mechanism and dynamically activate features in the model.

\subsection{Prompt Learning}

Prompt learning was first studied in the NLP field as a method for fine-tuning pre-trained language models (PLMs) to downstream tasks. This method adds some textual prompts to the input and helps PLM directly generate some required output text. 
The effectiveness of prompt learning and its advantage of only updating a small portion of parameters have recently attracted widespread attention. CoOp \cite{zhou2022learning} fine-tuning CLIP \cite{radford2021learning} by optimizing a set of continuous prompt vectors in its language branch for few-shot image recognition. CoCoOp \cite{zhou2022conditional} addresses the overfitting problem in CoOp and proposes a dynamic prompt based on visual features to improve the performance of generalization tasks. MaPLE \cite{khattak2023maple} proposed a multimodal prompt learning method that combines the visual and linguistic branches of CLIP to learn hierarchical prompts. 

To incorporate the prompt description information from different scenes and dynamically adjust network structures for images of varying complexities in different scene domains. We construct a gating module based on prompt learning, which enhances the representation power of the features and guides the learning of the gating module for different scenarios input.

%% file: 3_method.tex
\section{Methodology}
\subsection{Framework}

Given a source domain $\mathcal{D}^s=\lbrace (x^s_i,y^s_i) \rbrace ^{N_s} _{i=1}$ containing $\mathcal{{N_s}}$ samples.
Single-domain generalization aims to learn a model that can generalize to many unseen target domains $\mathcal{D}^t=\lbrace (x^t_i) \rbrace ^{N_t} _{i=1}$ using only the source domain data without prior knowledge about the target domains $\mathcal{D}^t$.
To improve the generalization ability of the model, we propose a prompt-based dynamic object-centric learning network for single-domain generalization as shown in Figure ~\ref{arch}. 
It contains two key components, the prompt-based object-centric gating module and the dynamic selective module. The prompt-based object-centric gating module fuses the text prompt embeddings with the visual features to learn enhanced scene information and extract object-centric representation from the fusion feature via Slot Attention. 
The dynamic selective module is used to dynamically activate the components of the network. With the gating masks output by the prompt-based object-centric gating module, we dynamically select feature maps from the blocks of the model backbone in both spatial and channel dimensions. In the spatial dimension, it identifies the spatial regions that contain significant object-centric information by the gating masks. Similarly, in the channel dimension, the gating masks help us select the most relevant channels that capture object-centric features.

The prompt learning module is based on the CLIP \cite{radford2021learning} which combines an image encoder and a text encoder and bridges the representation of visual and textual in joint space. We designed text description prompts in different image scenes and got the prompt embedding with the frozen text encoder of CLIP \cite{radford2021learning}. Guided by the prompts, the dynamic object-centric network can learn and extract more valuable information from the scene, and improve the performance in various scene-related tasks.

\begin{figure}[t]
\vspace{-0.5em}
\begin{center}
\includegraphics[width=1\linewidth]{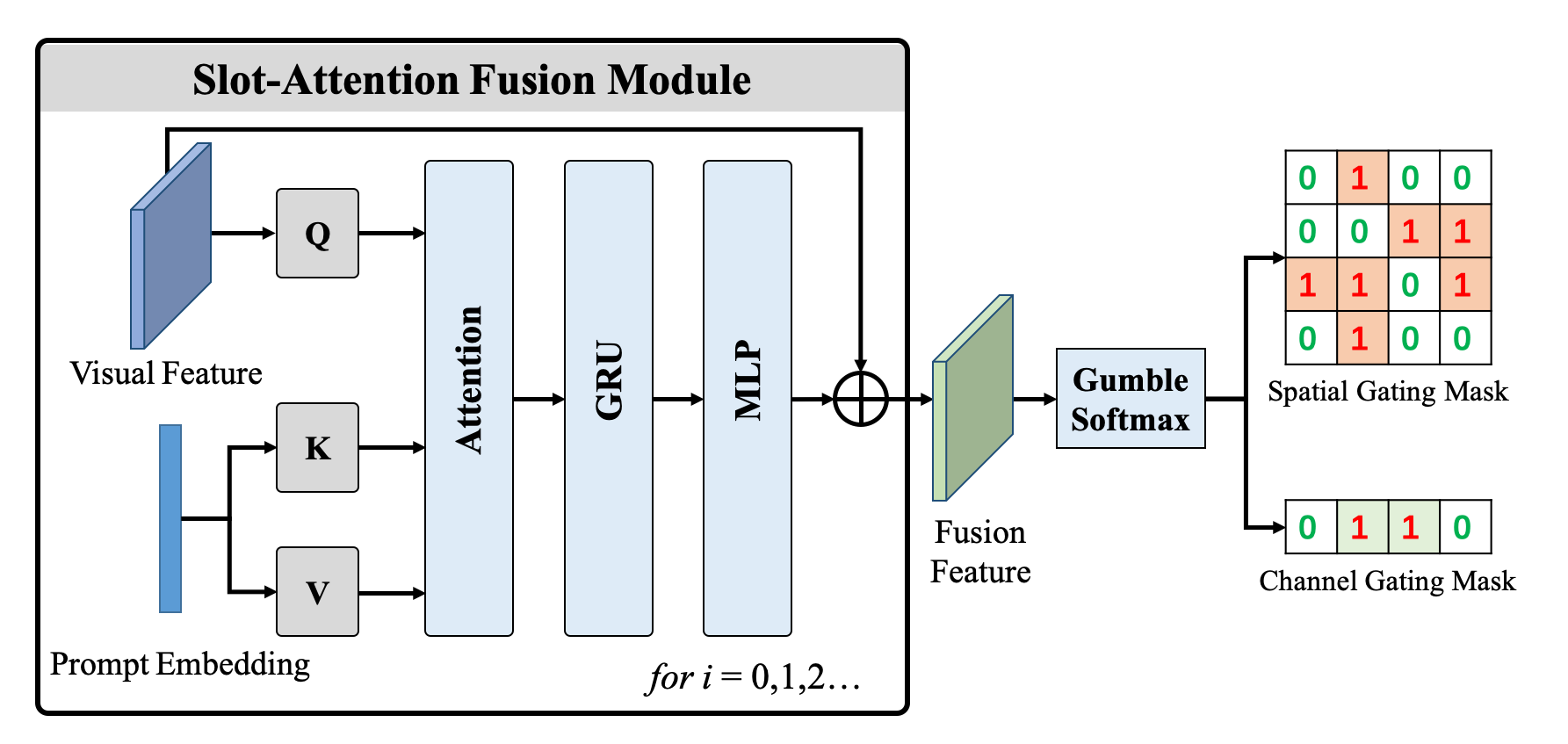}
\end{center}
\caption{Illustration of our proposed prompt-based object-centric gating module. 
   }
\label{gate_module}
\end{figure}

\subsection{Prompt-based Object-Centric Gating Module}

The dynamic network method based on gate function shows remarkable versatility and applicability and can be applied to different networks. Most works only utilize visual features for learning gating modules. These methods may involve biases in scene information and lead to overfitting problems to scenes, thereby hindering generalization capabilities to new scenes. 
To alleviate this issue, we utilize specifically designed scene prompts as compensation to obtain diverse information from various scenes. 
Object-centric representations can improve generalization capabilities by capturing essential visual attributes. By explicitly modeling objects, the learned representations can capture meaningful and transferable object-centric features that are robust to variations in appearance, context, or scene complexity.
In order to fully leverage both textual prompt information and visual features and then extract meaningful object-centric representation, we have developed a multimodal fusion module based on Slot Attention, as shown in Figure ~\ref{gate_module}.
Slot Attention is an attention mechanism that focuses attention on different slots, where each slot corresponds to a specific object or concept. 
We use the visual features as the initial slot and set the linearly transformed features as the Query $\mathcal{Q}$ in the Slot Attention mechanism. The linear transformations are applied to the prompt embedding to obtain the Key $\mathcal{K}$, and Value $\mathcal{V}$.
The attention score $\mathcal{A}$ is obtained by calculating the dot product between Query $\mathcal{Q}$ and Key $\mathcal{K}$ and followed it with softmax function:
\begin{equation}
\mathcal{A}=\operatorname{Softmax}\left(\frac{Q \cdot K^{\top}}{\sqrt{d_Q}}\right),
\end{equation}
where, ${\sqrt{d_Q}}$ is the dimension of Query $\mathcal{Q}$. Then, the attention features ${F_{att}}$ are obtained by the cross-product operation of attention score $\mathcal{A}$ and Value $\mathcal{V}$:
\begin{equation}
F_{att}=\mathcal{A} \cdot \mathcal{V}
\end{equation}
In addition, the slots are updated with loop iteration. During each iteration $t=1, \cdots, T$, we use the GRU function to update the features of each iteration. Based on Slot Attention, each prompt embedding is gradually refined according to relevant visual features.
This approach allows us to explicitly model and extract object-level feature representation. With the prompt embeddings and visual features, the Slot Attention aggregates the multi-modal features by weighting them based on the importance and relevance of the objects. 
The fused features are converted into gate functions:
\begin{equation}
{ slots }=\mathrm{GRU}({ state }={ slots, inputs={F_{att}} })
\end{equation}
Guided by the prompt embeddings, the Slot Attention fusion module can obtain the features that are relevant to the objects or concepts specified in the prompt. 
The gating function takes the fused features as input and generates gating masks. The gating masks act as a gate or filter that controls the flow of information within the model.
Since the gating function is a binary function that is not differentiable, during the training process, the Gumbel-Softmax technique is employed to transform the discrete binary function into a continuous variable.

\subsection{Dynamic Selective Module}

Based on the designed object-centric gating module, we embed the gating unit into the model to achieve dynamic activation of the model. Here, we take ResNet \cite{he2016deep} as an example and selectively activate connections from both spatial and channel levels to improve the generalization of the model. 
For channel-wise selective modules, we insert the selective module between the two convolutions of the block and dynamically select the feature information that should be input to the next layer. The binary mask output by the gate module is multiplied by the activation results of the convolutional layer to filter out the unimportant features. The binary mask can be expressed as follows:

\begin{equation}
M(i)= \begin{cases}1 & Slot(i) \geq threshold \\ 0 & \text { Otherwise }\end{cases},
\end{equation}
where, ${Slot_c(i)}$ is the feature of the ${i}$-th output by the slot attention multi-modal fusion module.

For the dynamic selective module, in each block of ResNet \cite{he2016deep}, the binary masks are obtained with the visual features and the prompt embedding through the above gate module. For the feature pyramid and the problem of different feature scales, we use the upsampling method to generate new gated features to adapt to the feature size of each layer. The masks are multiplied by the normalized features after convolution, thus filtering irrelevant spatial area features.
By dynamically activating features in the network at both spatial and channel levels, different levels of sparsity can be achieved in blocks. The dynamic object-centric perception approach prevents the model from overfitting and enhances the generalization ability on single-domain generalization tasks.

\subsection{Overall Training Objective}

To ensure stable training of the dynamic model, we adopt the approach proposed by Verelst et al. \cite{verelst2020dynamic} and introduce a bound loss to guide the model optimization. This bound loss constrains the sparsity of features in both spatial and channel dimensions, limiting it within the range of $\left[p \sqrt{T_d}, 1-p\left(1-\sqrt{T_d}\right)\right]$. Here ${T_d}$ denotes the target rate.
The lower and upper bounds of the regularization term can be expressed as: 
\begin{equation}
\begin{aligned}
& L_{b,\text { low }}=\sum_{l=1}^L \sum_{k \in\{s, c\}} \max \left(0, p \sqrt{T_d}-\left|M_k^l\right|_d\right)^2 \\
& L_{b,\text { up }}=\sum_{l=1}^L \sum_{k \in\{s, c\}} \max \left(0, p\left(1-\sqrt{T_d}\right)-1+\left|M_k^l\right|_d\right)^2
\end{aligned}
\end{equation}
where $|\cdot|_d$ is the density of the binary masks, and the exponential annealing function $p=\exp (-\alpha \cdot$ epoch $)$ is used to gradually loose the bound. We set the $\alpha$ to be 0.05 in our experiments.

By combining the loss function of the task and the bound loss function, the joint training loss function for our proposed method can be expressed as:

\begin{equation}
\mathcal{L}_{\mathrm{total}}=\mathcal{L}_{task}+\lambda_{b}(L_{b, \text { low }}+ L_{b, \text { up }}),
\label{total_loss}  
\end{equation}
where $ \lambda_{b} $ are the weight of the bound loss.

%% file: 4_experiments.tex
\section{Experiments}

To evaluate the effectiveness of our method, we conducted experiments on various visual task scenarios, such as image classification and object detection.
\subsection{Datasets}

\textbf{PACS} \cite{li2017deeper} is a generalization benchmark data set in the image classification domain, which contains four fields, namely art paintings, cartoons, photos, and sketches. Each domain contains 7 categories of images, a total of 9,991 images, and the image size is 224 × 224 pixels. This dataset has large stylistic differences between domains and is more challenging. For a fair comparison, we use the official split strategy to obtain the training set, validation set, and test set.

\textbf{Diverse-Weather Dataset.} We also evaluated our method on the urban-scene detection domain generalization benchmark diverse weather dataset built by \cite{wu2022single}. It contains five domains with different weather conditions, namely Daytime Clear, Night Clear, Dusk Rainy, Night Rainy, and Daytime Foggy. 
Here we use Daytime Clear data as the source domain and other domains as the target domain. The Daytime Clear domain consists of 19,395 training images, and 8,313 images are used as the validation set for model selection. 
The four other domains are set as target domains, including 26,158 images in the Night Clear scene, 3,501 images in the Dusk Rainy scene, 2,494 images in the Night Rainy scene, and 3,775 images in the Daytime Foggy scene.

\subsection{Image Classification}
\subsubsection{Implementation Details}
For the domain generalization task of image classification, we conducted evaluation experiments on single-source domain generalization and multiple-domain generalization on the PACS dataset. For single-source domain generalization experiments, four sets of experiments were conducted with one domain as the source domain and the others as the target domain. For multiple-domain generalization experiments, four sets of experiments were conducted with one of the four domains as the target domain and the other domains as the source domain. 
We have designed various prompts based on the template \textit{an image taken in \{scene name\}} for different scenarios.
ResNet-18 \cite{he2016deep} pre-trained on Imagenet is used as the backbone network of the model and fine-tuned on the source domain. The four-layer block of ResNet-18 integrates a prompt-based dynamic selective module to connect the features in the block at the spatial level and channel level. During the training process, we train the model in 70 epochs, the batch size is set to 256. We also set the learning optimizer as SGD with a weight decay of 0.0001, and the learning rate is 0.001.

\subsubsection{Experimental Results and Analysis}

\textbf{Single Domain Generalization.}Table~\ref{pacs_single_result} shows the experimental results of our single-domain generalization method on the PACS dataset. We compared our method with state-of-the-art methods such as ERM \cite{koltchinskii2011oracle}, RSC  \cite{huang2020self}, ASR \cite{fan2021adversarially}, and Meta-Casual \cite{chen2023meta}. Our method outperforms the state-of-the-art method with 1.2\% on average classification accuracy. Specifically, our method can boost the performance by 2.5\% than other methods in the cartoon domain with relative margins. The results verify the advantages of our proposed prompt-based dynamic object-centric learning method on single-domain generalization tasks.

\begin{table}[]
\small
\begin{center}
\caption{Single domain generalization image classification results (\%) on PACS with backbone of ResNet-18 \cite{he2016deep}. 
}
\label{pacs_single_result}
\begin{tabular}{l|cccc|c}
\hline
Method  &  A & C & S & P & Avg  \\ \hline
ERM  \cite{koltchinskii2011oracle} & 70.90 & 76.50 & 53.10 & 42.20 & 60.70 \\
RSC  \cite{huang2020self}& 73.40& 75.90 & 56.20 & 41.60 & 61.80 \\
RSC+ASR \cite{fan2021adversarially}& 76.70 & 79.30 & 61.60 & 54.60 & 68.10 \\
Meta-Casual \cite{chen2023meta} & 77.13 & 80.14 & 62.55&  59.60 & 69.86 \\
\hline             
Ours       & \textbf{78.77} &\textbf{82.69}&\textbf{62.94}&\textbf{60.09}&\textbf{71.12}   \\ 
\hline\end{tabular}
\end{center}
\end{table}

\textbf{Multiple Domain Generalization.} We also extended our method to multiple-domain generalization and conducted evaluation experiments on the PACS data set. The experimental results are shown in Table~\ref{pacs_multiple_result}. The split strategy of the training set, 
Existing multi-domain generalization methods can be categorized into two classes. The first type of methods requires domain recognition in the training stage, including DSN \cite{bousmalis2016domain}, Fusion \cite{mancini2018best}, MetaReg \cite{balaji2018metareg}, EpiFCR \cite{li2019episodic}, MASF \cite{dou2019domain} and DMG \cite{chattopadhyay2020learning}. The second category of methods does not utilize domain identity information in the training phase, consistent with a more realistic hybrid latent domain setting \cite{matsuura2020domain}. These methods include AGG \cite{li2019episodic}, HEX \cite{wang2019learning}, PAR \cite{wang2019learning}, 
ADA \cite{volpi2018generalizing}, ME-ADA \cite{zhao2020maximum}, MMLD \cite{matsuura2020domain} and Meta-Casual \cite{chen2023meta}. From Table~\ref{pacs_multiple_result} we can see that our method boosts the average classification accuracy with 1.0\% compared to the baseline methods. This result demonstrates the effectiveness of our proposed method on multi-domain generalization. 

\begin{figure}[]
\begin{center}
\includegraphics[width=0.8 \linewidth]{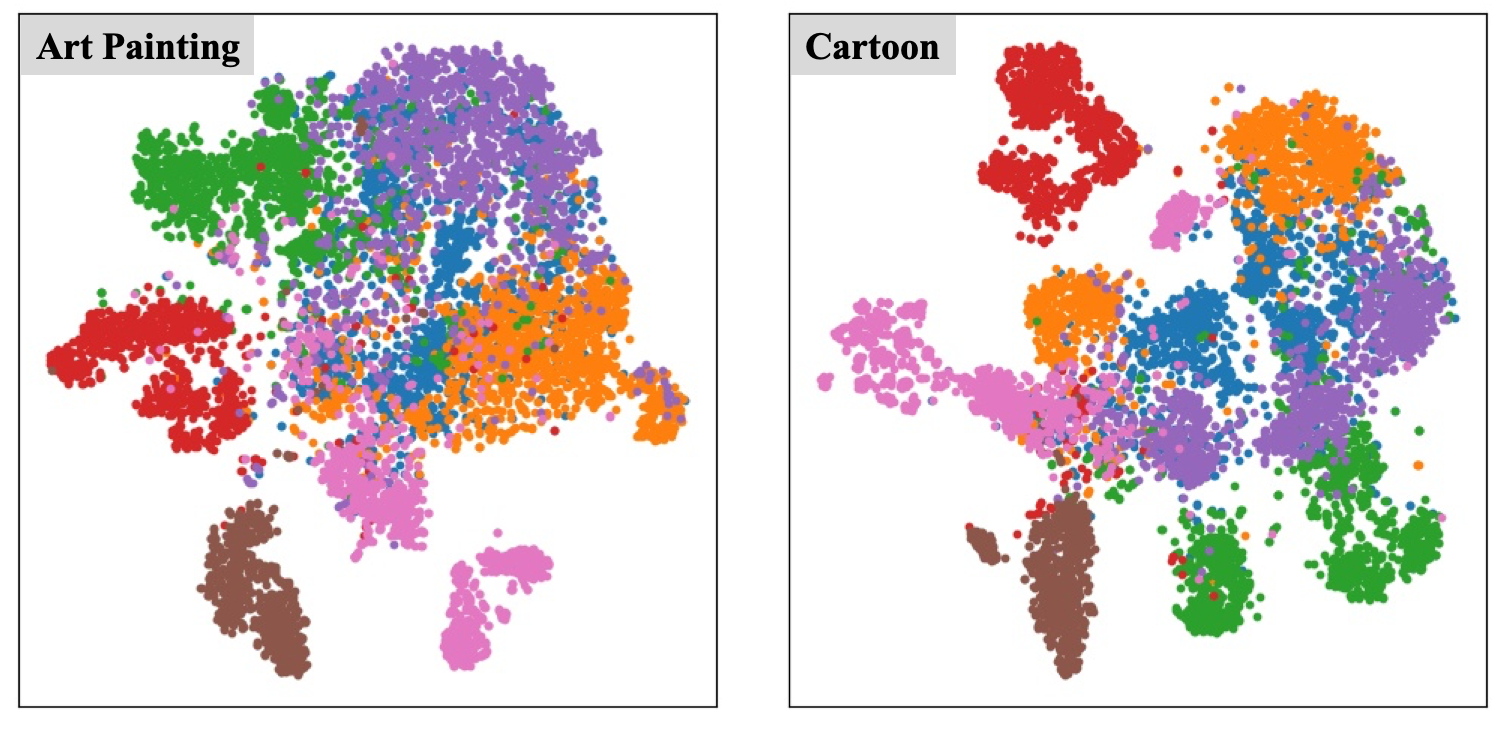}
\end{center}
\caption{The t-SNE of our method on the target domain of PACS. The upper left domain name is the source domain.
   }
\label{classify_vis}
\end{figure}

\textbf{Visualization Analysis.} We conducted a visualization analysis on the learned representations of image classification in Figure \ref{classify_vis}. From the visualization results, it can be seen that our method can effectively distinguish samples from the target domain in classification tasks.

\begin{table}[]
\small
\begin{center}
\caption{Multiple domain generalization image classification results (\%) on PACS with backbone of ResNet-18 \cite{he2016deep}. The domain name in the column is set as the target domain. 
}
\label{pacs_multiple_result}
\begin{tabular}{l|cccc|c}
\hline
Method  &  A  & C & P & S & Avg  \\ \hline
MetaReg  \cite{balaji2018metareg} & 83.70& 77.20 & 95.50& 70.30 & 81.70 \\
GUD \cite{volpi2018generalizing}& 78.32 &77.65 & 95.61 &74.21 & 81.44\\
Epi-FCR \cite{li2019episodic}& 82.10 &77.00 & 93.90& 73.00 & 81.50 \\
MASF \cite{dou2019domain}& 80.29 &77.17 & 94.99 &71.68 & 81.03 \\
DMG \cite{chattopadhyay2020learning}& 76.90& 80.38 & 93.55 &75.21 & 81.46 \\
DDAIG \cite{zhou2020deep}& 84.20 &78.10 & 95.30 &74.70 & 83.10 \\
CSD \cite{piratla2020efficient}& 78.90& 75.80 & 94.10& 76.70 & 81.40 \\
L2A-OT \cite{zhou2020learning}& 83.30& 78.20 & 96.20 &73.60 & 82.80 \\
EISNet \cite{wang2020learning}& 81.89& 76.44 & 95.93 &74.33 &82.15 \\
RSC \cite{huang2020self}& 83.43 &80.31 & 95.99& 80.85 &85.15 \\
ME-ADA \cite{zhao2020maximum}& 78.61& 78.65 & 95.57& 75.59 &82.10 \\
MMLD \cite{matsuura2020domain}& 81.28 &77.16 & 96.09& 72.29 &81.83 \\
L2D \cite{wang2021learning}& 81.44 &79.56 & 95.51& 80.58 &84.27 \\
FACT \cite{xu2021fourier}& 85.37& 78.38 & 95.15 &79.15 &84.51 \\
MatchDG \cite{mahajan2021domain}&81.32& 80.70 & 96.53 &79.72 &84.57 \\
CIRL \cite{lv2022causality}& 86.08& 80.59 & 95.93 &82.67 &86.32 \\
Meta-Casual \cite{chen2023meta} & 85.30 &80.93 & 96.53& 85.24 & 87.00 \\
\hline             
Ours       & \textbf{86.94} &\textbf{82.50}&\textbf{97.30}&\textbf{85.55}&\textbf{88.07}   \\ 
\hline\end{tabular}
\end{center}
\end{table}

\subsection{Object Detection}
\subsubsection{Implementation Details}

In order to further verify the effectiveness of our method, we also evaluate it on more complex target detection tasks. Compared with image classification tasks, target detection tasks not only require the correct classification of objects but also the accurate positioning of objects. Similar to other single-domain generalization methods for object detection, the Faster-RCNN \cite{ren2015faster} used in the experiment has the backbone of ResNet-101 \cite{he2016deep}. Here we conduct experiments on a dataset of traffic scenes. Following other object detection domain generalization methods, here we use the data of the Daytime Clear domain as the training set, and other domains are set as four target domains in the experiments. We train the model in 1000 iterators with a batch size of 4 images. The learning optimizer is SGD with a weight decay of 0.0005, and the learning rate is 0.001.

\subsubsection{Experimental Results and Analysis}

\textbf{Comparison with the State of the Art.} 
We compared with the state-of-the-art single-domain generalization object detection method Single-DGOD \cite{wu2022single} and Clip-Gap \cite{vidit2023clip} and the feature normalization domain generalization methods SW \cite{pan2019switchable}, IBN-Net \cite{pan2018two}, IterNorm \cite{huang2019iterative}, and ISW \cite{choi2021robustnet}. FasterRCNN \cite{ren2015faster}is a simple baseline method that initializes the parameters of the model through ImageNet pre-trained weights.
We set the Daytime Clear domain as the source domain and test the generalization performance on Daytime Foggy, Night Rainy, Dusk Rainy, and Night Clear four unseen target domains with more complex scenes. Table~\ref{dgod_result} shows the results of single-domain generalization for object detection.
It can be seen that, due to the domain offset, the test performance of all the methods on the target domain drops sharply. This phenomenon reflects the importance of model generalization performance.
Compared with the other methods, the performance of our method on the target domain is higher than that of the baseline method. Among them, there is a significant improvement in the Night Clear and Dusk Rainy domains, which are improved by 1.6\% and 1.4\% respectively. Our method improved by 0.6\% in the Daytime Foggy scene, and by 0.5\% in the challenging composite domain Night Rainy (contains two style changes at night and rainy). The experimental results demonstrate the effectiveness of our object-centric learning method in single-domain generalization for object detection. 

\begin{figure*}[t]
\begin{center}
\vspace{-0.5em}
\includegraphics[width=1 \linewidth]{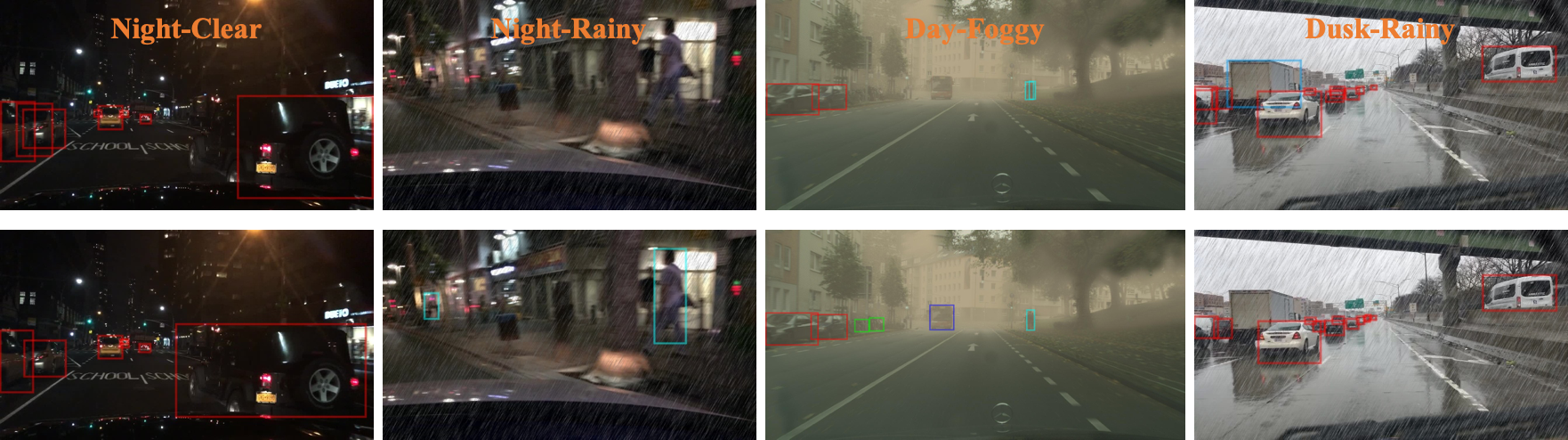}
\end{center}
\caption{Qualitative results of object detection on the urban scene Diverse-Weather Dataset. 
We visualized the detection results in the target domain, where the top row represents the detection results of CLIP-Gap \cite{vidit2023clip}, and the bottom row corresponds to our proposed method. It can be observed that in the "Night-Clear" scene, our method achieves more accurate car detection compared to CLIP-Gap \cite{vidit2023clip}. In the complex "Night-Rainy" scene, CLIP-Gap \cite{vidit2023clip} fails to detect the person, whereas our method successfully detects the person. In the "Day-Foggy" scene, our method accurately detects small-sized buses. Furthermore, in the "Dusk-Rainy" scene, our method exhibits improved accuracy in identifying and localizing trucks.
   }
\label{object_vis}
\end{figure*}

\textbf{Daytime Clear to Night Clear.} 
Table~\ref{foggy_result} shows the detection results on the Night Clear scene. Compared to the daytime scenes in the source domain, nighttime scenes pose challenges for object recognition and detection due to low visibility conditions. 
From the experimental results, it can be observed that our method outperforms other methods in various object categories. Specifically, the performance on bus, motor, and person categories has been improved by 3.2\%, 2.1\%, and 2.8\% respectively. These results demonstrate the effective generalization ability of our dynamic network method to Daytime Clear to Night Clear scenarios.

\begin{table}[]
\vspace{-0.5em}
\scriptsize
\begin{center}
\caption{Single domain generalization object detection results (\%).
}
\label{dgod_result}
\begin{tabular}{l|c|cccc}
\hline
Method  & \multicolumn{1}{c|}{\begin{tabular}[c]{@{}c@{}}Day \\ Clear\end{tabular}}  &\multicolumn{1}{c}{\begin{tabular}[c]{@{}c@{}}Night \\ Clear\end{tabular}}   &\multicolumn{1}{c}{\begin{tabular}[c]{@{}c@{}}Dusk\\ Rainy\end{tabular}}  &\multicolumn{1}{c}{\begin{tabular}[c]{@{}c@{}}Night\\ Rainy \end{tabular}} &\multicolumn{1}{c}{\begin{tabular}[c]{@{}c@{}}Day\\ Foggy\end{tabular}}\\ \hline
Faster-RCNN  \cite{ren2015faster} & 48.1 & 34.4 & 26.0 & 12.4 & 32.0\\
IterNorm  \cite{huang2019iterative}& 43.9& 29.6 & 22.8 & 12.6 &28.4\\
SW  \cite{pan2019switchable}& 50.6 & 33.4 & 26.3 & 13.7 & 30.8\\
IBN-Net \cite{pan2018two} & 49.7 & 32.1 & 26.1 & 14.3 & 29.6\\
ISW \cite{choi2021robustnet} & 51.3 & 33.2 & 25.9 & 14.1 & 31.8\\
S-DGOD \cite{wu2022single} & \textbf{56.1} & 36.6 & 28.2& 16.6 &33.5\\
CLIP-Gap \cite{vidit2023clip} & 51.3 & 36.9& 32.3& 18.7 &38.5\\
\hline             
Ours       &53.6 &\textbf{38.5}&\textbf{33.7}&\textbf{19.2}&\textbf{39.1}   \\ 
\hline\end{tabular}
\end{center}
\end{table}

\begin{table}[]
\vspace{-0.5em}
\scriptsize
\setlength{\tabcolsep}{3pt}
\begin{center}
\caption{Per-class results(\%) on Day-Clear to Night-Clear. 
}
\label{night_sunny_result}
\begin{tabular}{l|ccccccc|c}
\hline
Method  & bus  & bike & car & motor  & person    & rider       & truck & mAP       \\ \hline
Faster-RCNN  \cite{ren2015faster} & 34.7 &32.0 &56.6 &13.6 &37.4 &27.6 &38.6 &34.4\\
IterNorm  \cite{huang2019iterative}& 38.5& 23.5 &38.9 &15.8 &26.6 &25.9 &38.1 &29.6\\
SW  \cite{pan2019switchable}& 38.7& 29.2 &49.8 &16.6 &31.5 &28.0 &40.2 &33.4\\
IBN-Net \cite{pan2018two} & 37.8 & 27.3 & 49.6 & 15.1&  29.2 & 27.1 & 38.9 &32.1\\
ISW \cite{choi2021robustnet} & 38.5&  28.5 & 49.6 & 15.4&  31.9&  27.5&  41.3 & 33.2\\
S-DGOD \cite{wu2022single} & 40.6& 35.1 &50.7 &19.7 &34.7 &32.1 &43.4 &36.6\\
CLIP-Gap \cite{vidit2023clip} & 37.7&  34.3&  58.0&  19.2&  37.6&  28.5&  42.9&  36.9\\
\hline             
Ours       &\textbf{40.9}&\textbf{35.0}&\textbf{59.0}& \textbf{21.3}      &\textbf{40.4}&\textbf{29.9} &\textbf{42.9} &\textbf{38.5}                      \\ 
\hline\end{tabular}
\end{center}
\end{table}

\textbf{Daytime Clear to Dusk Rainy.} 
Table~\ref{dusk_rainy_result} shows the detection results on the Dusk Rainy scene. This scene is affected by low light conditions and rain and has a large domain shift from the source daytime image. Compared with other methods, our method has comparable performance on various categories of objects. Particularly, our method improves about 2.3\%, 3.6\%, and 3.1\% on the bus, motor, and person categories, respectively. This shows that our dynamic network method can effectively improve the generalization performance of the model from Daytime Clear to Dusk Rainy.

\begin{table}[]
\vspace{-0.5em}
\scriptsize
\setlength{\tabcolsep}{3pt}
\begin{center}
\caption{Per-class results(\%) on Day-Clear to Dusk-Rainy.
}
\label{dusk_rainy_result}
\begin{tabular}{l|ccccccc|c}
\hline
Method  & bus  & bike & car & motor  & person    & rider       & truck & mAP       \\ \hline
Faster-RCNN  \cite{ren2015faster} & 28.5 &20.3 &58.2 &6.5 &23.4 &11.3 &33.9 &26.0\\
IterNorm  \cite{huang2019iterative}& 32.9 &14.1 &38.9 &11.0 &15.5 &11.6 &35.7 &22.8\\
SW  \cite{pan2019switchable}& 35.2 &16.7 &50.1 &10.4 &20.1 &13.0 &38.8 &26.3\\
IBN-Net \cite{pan2018two} & 37.0 &14.8 &50.3 &11.4 &17.3 &13.3 &38.4 &26.1\\
ISW \cite{choi2021robustnet} & 34.7 &16.0 &50.0 &11.1 &17.8 &12.6 &38.8 &25.9\\
S-DGOD \cite{wu2022single}& 37.1 &19.6 &50.9 &13.4 &19.7 &16.3 &40.7 &28.2\\
CLIP-Gap \cite{vidit2023clip} & 37.8 &22.8 &60.7&16.8&26.8 &\textbf{18.7}&42.4&32.3\\
\hline             
Ours       &\textbf{39.4}&\textbf{25.2}&\textbf{60.9}&\textbf{ 20.4}    &\textbf{29.9}&16.5 &\textbf{43.9} &\textbf{33.7}                      \\ 

\hline\end{tabular}
\end{center}
\end{table}

\textbf{Daytime Clear to Night Rainy.} 
Table~\ref{night_sunny_result} shows the results on the Dusk Rainy scene. The nighttime rainy scene contains the effects of both low-light and rainy weather environments, and there is a large domain shift from the source daytime image. The influence of this composite domain shift brings huge challenges to object detection, which leads the model to suffer serious performance degradation. Compared with other methods, our method improves the average mAP by 0.5\% and improves in the person and rider categories by 1.9\% and 3.3\%, respectively. The effectiveness of our method for challenging target domain scenarios is further verified.

\begin{table}[]
\vspace{-0.5em}
\scriptsize
\setlength{\tabcolsep}{3pt}
\begin{center}
\caption{Per-class results(\%) on Day-Clear to Night-Rainy. 
}
\label{night_rainy_result}
\begin{tabular}{l|ccccccc|c}
\hline
Method  & bus  & bike & car & motor  & person    & rider       & truck & mAP       \\ \hline
Faster-RCNN  \cite{ren2015faster} & 16.8 & 6.9& 26.3& 0.6& 11.6& 9.4& 15.4 &12.4 \\
IterNorm  \cite{huang2019iterative}& 21.4& 6.7& 22.0& 0.9& 9.1& 10.6& 17.6& 12.6\\
SW  \cite{pan2019switchable}& 22.3& 7.8& 27.6& 0.2& 10.3& 10.0&17.7& 13.7\\
IBN-Net \cite{pan2018two} & 24.6& 10.0&28.4&0.9& 8.3& 9.8 &18.1& 14.3\\
ISW \cite{choi2021robustnet} & 22.5 &11.4 &26.9 &0.4& 9.9 &9.8 &17.5& 14.1\\
S-DGOD \cite{wu2022single} & 24.4& 11.6& 29.5 &9.8 &10.5 &11.4& 19.2 &16.6\\
CLIP-Gap \cite{vidit2023clip} & \textbf{28.6} &12.1& \textbf{36.1}& 9.2& 12.3& 9.6 &22.9& 18.7\\
\hline             
Ours       &25.6&\textbf{12.1}&35.8&\textbf{ 10.1}      &\textbf{14.2}&\textbf{12.9} &\textbf{22.9} &\textbf{19.2}                      \\ 
\hline\end{tabular}
\end{center}
\end{table}

\textbf{Daytime Clear to Day Foggy.} 
Table~\ref{night_rainy_result} shows the detection results on the Day Foggy scene. Objects in foggy scene images are blurred, which brings challenges to object detection. Our method shows comparable performance on various categories of objects in this scene. This shows that our dynamic network method can effectively improve the generalization performance of the model.

\textbf{Visualization Analysis.} We also conducted a visualization analysis on object detection as shown in Figure \ref{object_vis}. From the visualization results, it can be seen that our method can effectively extract object-centric features to accurately classify and locate objects in complex urban scenes. 

\begin{table}[]
\vspace{-0.5em}
\scriptsize
\setlength{\tabcolsep}{3pt}
\begin{center}
\caption{Per-class results(\%) on Day-Clear to Day-Foggy. 
}
\label{foggy_result}
\begin{tabular}{l|ccccccc|c}
\hline
Method  & bus  & bike & car & motor  & person    & rider       & truck & mAP       \\ \hline
Faster-RCNN  \cite{ren2015faster} & 28.1 &29.7 &49.7 &26.3 &33.2 &35.5 &21.5 &32.0
\\
IterNorm  \cite{huang2019iterative}& 29.7 &21.8 &42.4 &24.4 &26.0 &33.3 &21.6 &28.4\\
SW  \cite{pan2019switchable}& 30.6 &26.2 &44.6 &25.1 &30.7 &34.6 &23.6 &30.8\\
IBN-Net \cite{pan2018two} & 29.9 &26.1 &44.5 &24.4 &26.2 &33.5 &22.4 &29.6\\
ISW \cite{choi2021robustnet} & 29.5 &26.4 &49.2 &27.9 &30.7 &34.8 &24.0 &31.8\\
S-DGOD \cite{wu2022single} & 32.9 &28.0 &48.8 &29.8 &32.5 &38.2 &24.1 &33.5\\
CLIP-Gap \cite{vidit2023clip} & 36.1 &34.3&58.0&33.1&39.0&43.9&25.1 & 38.5\\
\hline             
Ours       &\textbf{36.1}&\textbf{34.5}&\textbf{58.4}& \textbf{33.3}     &\textbf{40.5}&\textbf{44.2} &\textbf{26.2} &\textbf{39.1}                      \\ 

\hline\end{tabular}
\end{center}
\end{table}

\subsection{Ablation Study}
To analyze the impact of different components in our proposed method, we conducted some ablation studies. First, we performed an ablation study by removing the prompt-based adaptation mechanism from our dynamic network approach. This analysis aimed to assess the significance of prompts in guiding the network dynamic adjustments. Second, we also conducted an additional ablation analysis to assess the contribution of the slot attention mechanism by replacing it with a traditional attention method. Table~\ref{abation_study} shows the results of the ablation experiment. It can be seen that when using dynamic networks for training, the performance is significantly improved, with an average accuracy increase of 9.3\%. When using the visual-language pre-trained CLIP \cite{radford2021learning} model and feature fusion based on slot attention, the generalization performance of our method is further improved, with an average accuracy increase of 6.8\%.

\begin{table}[]
\vspace{-0.5em}
\scriptsize
\setlength{\tabcolsep}{3pt}
\begin{center}
\caption{Ablation study (\%) on PACS dataset with backbone of ResNet-18 \cite{he2016deep}. The domain name in the column is used as the source domain, and the other domains are used as the target domains.
}
\label{abation_study}
\begin{tabular}{l|cc|cccc|c}
\hline
Method  & Dynamic & Attention &  Artpaint  & Cartoon  & Sketch & Photo  & Avg  \\ \hline
Base  &$\usym{2715}$&$\usym{2715}$& 71.26 & 67.64 & 43.97 & 36.99 & 54.97 \\ \hline    
Ours  & $\usym{1F5F8}$&$\usym{2715}$ & 74.29 & 78.54 & 56.54 & 47.74 & 64.27 \\
Ours  &$\usym{1F5F8}$& Normal & 75.78 & 81.94 & 59.94 & 58.09 &68.94  \\ 
Ours       & $\usym{1F5F8}$& Slot  & \textbf{78.77} &\textbf{82.69}&\textbf{62.94}&\textbf{60.09}&\textbf{71.12}    \\ 
\hline\end{tabular}
\end{center}
\end{table}

\section{Conclusion}
Due to the domain shift, models trained on a single domain often suffer from significant performance degradation when tested on unseen target domains.
Furthermore, different image scenes in real-world scenarios require varying model complexities, and static networks are prone to overfitting. In this paper, we propose a dynamic object-centric learning approach via prompts to dynamically adjust the network to perceive object-centric features, thereby enhancing the generalization performance. 
First, we propose a multimodal fusion module based on Slot Attention to extract object-centric features from objects. 
In addition, a prompt-based object-centric gating module is introduced to leverage the various scene prompts to guide the learning of the gating masks for various scenes. 
Finally, the object-centric gating masks are used to dynamically select the relevant object-centric feature within a model leading to more accurate and robust predictions.
Extensive experimental results conducted on image classification and object detection tasks validate the effectiveness and generalizability of our proposed method.

\section*{Acknowledgments}
This work was supported in part by the NSFC (under Grant 62376186, 61932009) and in part by the CAAI-Huawei MindSpore Open Fund.